\pdfoutput=1

\documentclass[11pt]{article}

\usepackage[hyperref]{acl}

\usepackage{times}
\usepackage{latexsym}
\usepackage{booktabs}

\usepackage{dirtytalk}
\usepackage{url}
\usepackage{hyperref}
\usepackage{amsmath,amssymb
}\usepackage{dsfont}

\usepackage{algorithm}
\usepackage{algpseudocode}
\usepackage{graphicx}

\usepackage{multirow}
\usepackage{array}
\usepackage{tikz}
\usetikzlibrary{fit}
\usetikzlibrary{shapes.geometric}
\usepackage{cleveref} 

\newcolumntype{H}{>{\setbox0=\hbox\bgroup}c<{\egroup}@{}}

\usepackage{xcolor}

\usepackage{bm}

\usepackage[T1]{fontenc}
\usepackage[utf8]{inputenc}

\usepackage{microtype}

\title{Probing Subphonemes in Morphology Models}

\author{Gal Astrach \quad \quad \quad Yuval Pinter \\
  Department of Computer Science and 
  Data Science Research Center \\
  Ben-Gurion University of the Negev\\
  Beer Sheva, Israel \\
  \texttt{\{galastra@post,uvp@cs\}.bgu.ac.il} \\}

\begin{document}
\maketitle
\begin{abstract}
    Transformers have achieved state-of-the-art performance in morphological inflection tasks, yet their ability to generalize across languages and morphological rules remains limited.
    One possible explanation for this behavior can be the degree to which these models are able to capture implicit phenomena at the phonological and subphonemic levels.
    We introduce a language-agnostic probing method to investigate phonological feature encoding in transformers trained directly on phonemes, and perform it across seven morphologically diverse languages.
    We show that phonological features which are local, such as final-obstruent devoicing in Turkish, are captured well in phoneme embeddings, whereas long-distance dependencies like vowel harmony are better represented in the transformer's encoder.
    Finally, we discuss how these findings inform empirical strategies for training morphological models, particularly regarding the role of subphonemic feature acquisition.
\end{abstract}

\section{Introduction}
\label{sec:intro}
The transformer architecture has revolutionized natural language processing and computational linguistics since its introduction by \citet{NIPS2017_3f5ee243}.
While it has achieved state-of-the-art results across various tasks, much remains to be understood about its inner workings and representations. Investigating how transformers acquire and use linguistic knowledge is crucial for assessing their ability to generalize beyond shallow pattern recognition.
One such aspect is morphological knowledge, such as that examined in tasks like \textit{morphological inflection}~\cite{cotterell-etal-2017-conll}, where a model predicts a word's inflected form given a lemma and morphosyntactic attributes.
For example, for the English lemma \say{hug} and the morphosyntactic attributes \texttt{VERB;PAST}, the model should output \say{hugged}.
In many languages, morphology interacts in meaningful ways with \emph{phonological} attributes, for example through the phenomenon of \emph{harmony}, raising the question of how this correspondence is manifested in model representations.
This question is hard to pursue due to the general scarcity of multilingual morphological data, which hinders models' ability to generalize to new lemmas and morphosyntactic attributes~\cite{goldman-etal-2022-un,kodner-etal-2023-morphological} and to adapt to the diversity of morphological processes~\cite{kodner-etal-2022-sigmorphon}.
In this work, we present a language-agnostic method for testing phonological features and long-context feature agreement in models trained on a morphological task.\footnote{\url{https://github.com/MeLeLBGU/probing-subphonemes}} We do so by training designated probing classifiers that predict linguistic properties from a model's internal representations~\cite{belinkov-2022-probing}.
We show how a morphological transformer implicitly acquires phonological knowledge, complementing previous findings regarding the representations found in neural phoneme embeddings~\cite{rodd-1997-recurrent,silfverberg-etal-2021-rnn,muradoglu-hulden-2023-transformer,mirea-bicknell-2019-using,silfverberg-etal-2018-sound,kolachina-magyar-2019-phone,steuer-etal-2023-information} and the information conveyed by morphological models~\cite{muradoglu-hulden-2023-transformer,kodner-etal-2023-exploring,gorman-etal-2019-weird}.
Unlike previous work, we demonstrate via explainability methods that model representations explicitly encode phonological features, and quantify how well they are encoded across languages and features.
We test the hypothesis that when trained on reliable phonological representations, models acquire subphonemic features such as \texttt{VOICE} or \texttt{ROUND}~\cite{chomsky1968sound} that play a role in morphology, and that this ability depends on a language's reliance on such features in encoding inflectional properties.
We find that local phenomena, 
such as final consonant devoicing, are captured in the character embeddings, while long-distance phenomena are better represented in contextualized embeddings from the transformer encoder.
Finally, we argue for current practices in language transfer of morphological models based on our results.

\section{Probing Phonological Features}\label{sec:method}
Our experiments consist of three stages: training a phonemic transformer model on a morphological task for a language; probing the embeddings for phonological features; and analyzing the probe using minimum description length~(MDL).

\subsection{Phoneme-based Transformer}
We use a character-based encoder-decoder transformer which achieves state-of-the-art results on morphological inflection~\cite{wu-etal-2021-applying}.
This architecture is relatively small and employs a feature-invariant positional encoding for morphosyntactic tags, making their order irrelevant to the model.
We modify the architecture by weight tying, using the same embedding table for both the encoder and decoder during training and evaluation~\cite{press-wolf-2017-using}.
We train the transformer on the SIGMORPHON 2017 shared task dataset~\cite{cotterell-etal-2017-conll} which covers multiple languages with diverse typologies.\footnote{For Hebrew (vocalized), we use the dataset from \citet{kodner-etal-2022-sigmorphon}.}
To directly analyze the representation of phonological features, we transcribe the lemmas from standard orthographic form to International Phonetic Alphabet~(IPA) using Epitran~\cite{mortensen-etal-2018-epitran},\footnote{Version 1.25.1.} a rule-based grapheme-to-phoneme tool.\footnote{For Hebrew (voc), we use a dedicated API~\citep{zemereshet}.}
We refer to IPA characters as \emph{phonemes} interchangeably.

We train two versions of the transformer: (i) an \textbf{inflection model}, trained on the phonemic transcriptions of the morphological inflection task; and (ii) a \textbf{lemma copying model}, where we replace each morphosyntactic attribute with \texttt{COPY} and set the inflected form as identical to the lemma.
We then probe the phoneme embeddings and the encoder using a set of probe tasks.

\subsection{Probe Tasks}
We design two types of probes to evaluate how well phonological features are embedded by a model.
The \textbf{phoneme probe} assesses the phoneme embeddings, while the \textbf{harmony probe} evaluates the encoder's output vectors.
Separate probes are trained for each (phonological feature, language) pair.

\paragraph{Extracting phonological features.} We use PanPhon~\cite{mortensen-etal-2016-panphon}\footnote{Version 0.21.1.} to map each phoneme to its corresponding phonological features, each represented as a ternary value: $+$, $-$, or $0$ (meaning \say{irrelevant}), demonstrated in \autoref{tab:panphon}.

\begin{table*}[t]
    \centering
    \resizebox{\linewidth}{!}{ 
    \begin{tabular}{l*{21}{c}}
        \toprule
        & syl & son & cons & cont & delrel & lat & nas & strid & voi & sg & cg & ant & cor & distr & lab & hi & lo & back & round & tense & long \\
        \midrule
        \texttt{k} (phoneme) & $-$ & $-$ & $+$ & $-$ & $-$ & $-$ & $-$ & $0$ & $-$ & $-$ & $-$ & $-$ & $-$ & $0$ & $-$ & $+$ & $-$ & $+$ & $-$ & $0$ & $-$ \\
        \texttt{köpek} (v. harmony) & $+$ & $+$ & $-$ & $+$ & $-$ & $-$ & $-$ & $0$ & $+$ & $-$ & $-$ & $0$ & $-$ & $0$ & $-$ & $-$ & $-$ & $-$ & $0$ & $0$ & $-$  \\
        \texttt{köpek} (c. harmony) & $-$ & $-$ & $+$ & $-$ & $-$ & $-$ & $-$ & $0$ & $-$ & $-$ & $-$ & $0$ & $-$ & $0$ & $0$ & $0$ & $-$ & $0$ & $-$ & $0$ & $-$  \\
        \bottomrule
    \end{tabular}
    }
    \caption{Phonology features extracted via Panphon for the probes: a single phoneme and (vowel / consonant) harmony type for the word \texttt{köpek} (dog in Turkish).}
    \label{tab:panphon}
\end{table*}

\begin{figure}[!t]
    \centering
    \includegraphics[width=0.5\textwidth]{{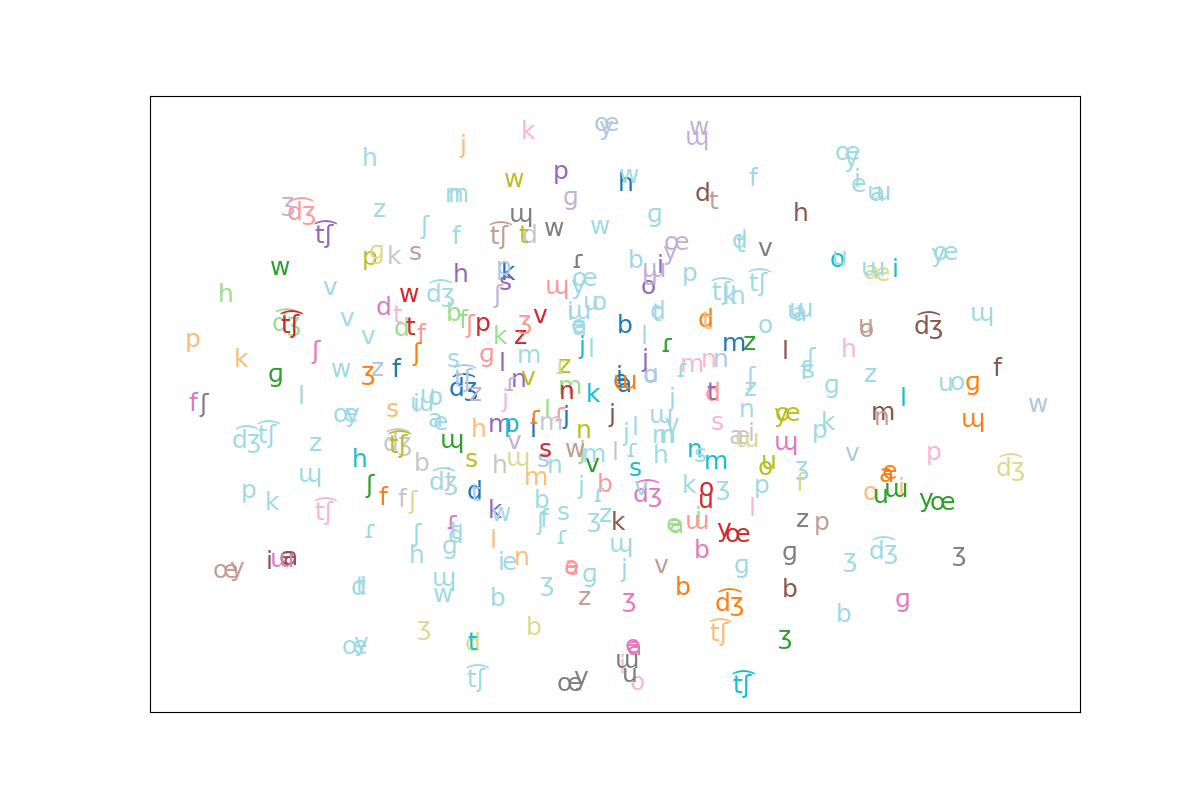}}
    \caption{t-SNE projection of phoneme embeddings after training on Turkish morphological inflection. Characters from each seed are presented in a distinct color.}
    \label{fig:tsne_turkish}
\end{figure}

\paragraph{Phoneme probe.} \label{par:phonemes_probe}
For each phonological feature, we train a probe using the phoneme embeddings as input and the feature values as labels. 
However, the limited number of phonemes per language makes it insufficient for training a probe and may leave some feature values out of distribution.
To mitigate this, we augment the dataset by training the transformer with multiple random seeds, generating a diverse set of phoneme embeddings. Due to computational constraints, we additionally apply oversampling by a factor of three.
To see whether embeddings from different seeds exhibit inherent structure, we project them into a 2D plane (\autoref{fig:tsne_turkish}) using t-SNE~\cite{van2008visualizing}.
The lack of clustering among identical phonemes suggests that this data augmentation strategy effectively diversifies the embeddings. 

\paragraph{Harmony probe.} \label{par:harmony_probe}
To investigate how well the transformer encodes long-distance phonological dependencies, we design a probe that mimics vowel and consonant harmony rules.
We generate nonce words using the method and code from \citet[\S3]{muradoglu-hulden-2023-transformer}.
The probe's inputs are the
contextualized phoneme vectors produced from the
encoder for these nonce words, taken from the last layer of the encoder when passed on nonce words, with beginning-of-sequence and end-of-sequence tokens attached.
The probe classifies the harmony type of each word for both vowel and consonant harmony: $+$ if all phonemes are $+$ or 0, $-$ if all are $-$ or 0, and 0 if the word is disharmonic, containing both $+$ and $-$ values, demonstrated in \autoref{tab:panphon}.
We train separate probes for vowel and consonant harmony for each phonological feature, but only if at least two phonemes in the language exhibit $+$ and $-$ values each for that feature.

\subsection{MDL Probes}
Traditional probing methods use metrics like accuracy or F1 score to estimate how well embeddings encode linguistic properties. However, this approach has several limitations.
A probe may perform well even with randomly assigned labels or when applied to randomly-initialized representations.
To address this, we adopt an information-theoretic approach~\citep{voita-titov-2020-information} and report a metric based on the probe's minimum description length (MDL) instead.
This method accounts for the probe's complexity, making it more robust and comparable across different models and linguistic properties.
For each phonological feature, we compute MDL using the online coding approach:
We segment the probe dataset at sequential indices $t_0 < t_1 < \cdots < t_S=n$, where $n$ is the size of the probe's dataset.
A probe $\theta_i$ is trained on each prefix of the data while measuring cross-entropy loss on the next segment.\footnote{To address class imbalance, we weigh the loss by the inverse frequency of each feature. 
The probe's architecture follows \citet{voita-titov-2020-information} and is implemented as a multi-layer perceptron with two hidden layers of $100$ neurons each.}
Summing these losses yields the total description length of the feature:
\begin{equation}
\small
L =t_0\log_2K-\sum_{i=0}^{S-1}\log_2p_{\theta_i}(y_{t_i + 1:t_{i+1}} | x_{t_i + 1:t_{i+1}}),
\label{eqn:compression1}
\end{equation}
where $K$ is the number of classes (in our case, 3).

To normalize across datasets of different sizes, we compute a \emph{compression score} by dividing the uniform coding length by the MDL, providing a comparable measure of how efficiently phonological information is encoded across different languages with varying phoneme inventories:

\begin{equation}
\mathcal{C} = \frac{n \log_2K}{L}.
\label{eqn:compression2}
\end{equation}

Analyze a probe using compression score characterizes the strength of regularity in the embeddings with respect to the labels. While it is a strictly relative metric, higher values indicate stronger regularity and therefore that the labels are better encoded in the embeddings.

\section{Results}
\label{sec:results}

\begin{figure*}[!ht]
    \centering
    \includegraphics[width=0.9\textwidth]{{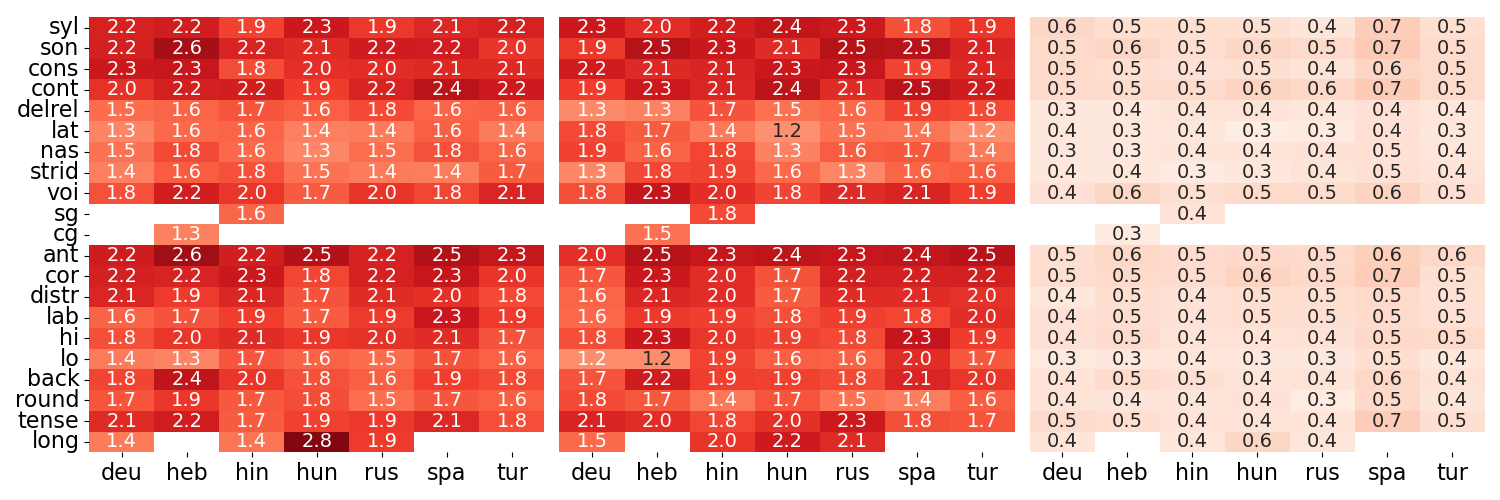}}
    \caption{Compression scores ($\mathcal{C}$) of phoneme embeddings: phonological features are plotted on the y-axis (abbreviated), and languages are on the x-axis (represented by their ISO 639-3 code). From left to right: inflection model, lemma copying, and control task.}
    \label{fig:inflection_experiment}
\end{figure*}

We apply our method to a set of seven languages selected to provide reasonable diversity with respect to morphophonological phenomena.\footnote{Feature inclusion score~\citep{ploeger-etal-2024-typological} of 0.63.}
These languages represent different morphological typologies: agglutinative / fusional, prefixing / suffixing / non-concatenative; and exhibit diverse phonological rules, such as palatalization and vowel harmony.
We focus on languages with low \emph{orthographic depth}, which allows us to probe phoneme embeddings in a way comparable to probing character embeddings in standard orthography.

We first compare the results to a control probing task 
where the phonological features (labels) are randomly shuffled \citep{hewitt-liang-2019-designing}.
The results, presented in \autoref{fig:inflection_experiment}, validate that the compression score is a good indicator of phonological feature representation in the embeddings, as all scores remain below 1.0.
We next discuss specific morphophonological phenomena and how they manifest in the data.

\paragraph{Turkish final-obstruent devoicing.}
\label{subsec:probing_phoneme}
In Turkish, a word-final \texttt{[-CONTINUANT]} consonant is devoiced.
Moreover, in accusative case the consonant becomes voiced, and a \texttt{[+HIGH, -ROUNDED]} vowel is suffixed, with the \texttt{BACK} value subject to vowel harmony.
For example, \textit{kebap} (the kebab) becomes \textit{kebabı} (\texttt{ACC} the kebab).
Our probe shows that both \texttt{VOICE} and \texttt{CONTINUANT} features have a relatively good compression score in both probes, with more prominence compared to other features in the inflection task.

\paragraph{Hungarian gemination.}
Nearly every phoneme in Hungarian has a corresponding \texttt{[+LONG]} variant of it in the phonetic inventory.
There are two processes that can alter the value of the \texttt{LONG} feature: gemination, where consonants at the end of a verb become \texttt{[+LONG]} before a suffix, and degemination, where \texttt{[+LONG]} consonants become \texttt{[-LONG]} when preceded or followed by another consonant. Among all languages and features in the phoneme probe, the compression score for \texttt{LONG} in Hungarian is the highest in the inflection model. We hypothesize that this is due to two factors: (i) morphological alternations affecting the gemination process, and (ii) the high entropy of \texttt{LONG}, which effectively separates Hungarian’s phonetic inventory, allowing the probe to achieve a relatively high compression score.

\begin{figure}[!t]
    \centering
    \includegraphics[width=0.4\textwidth]{{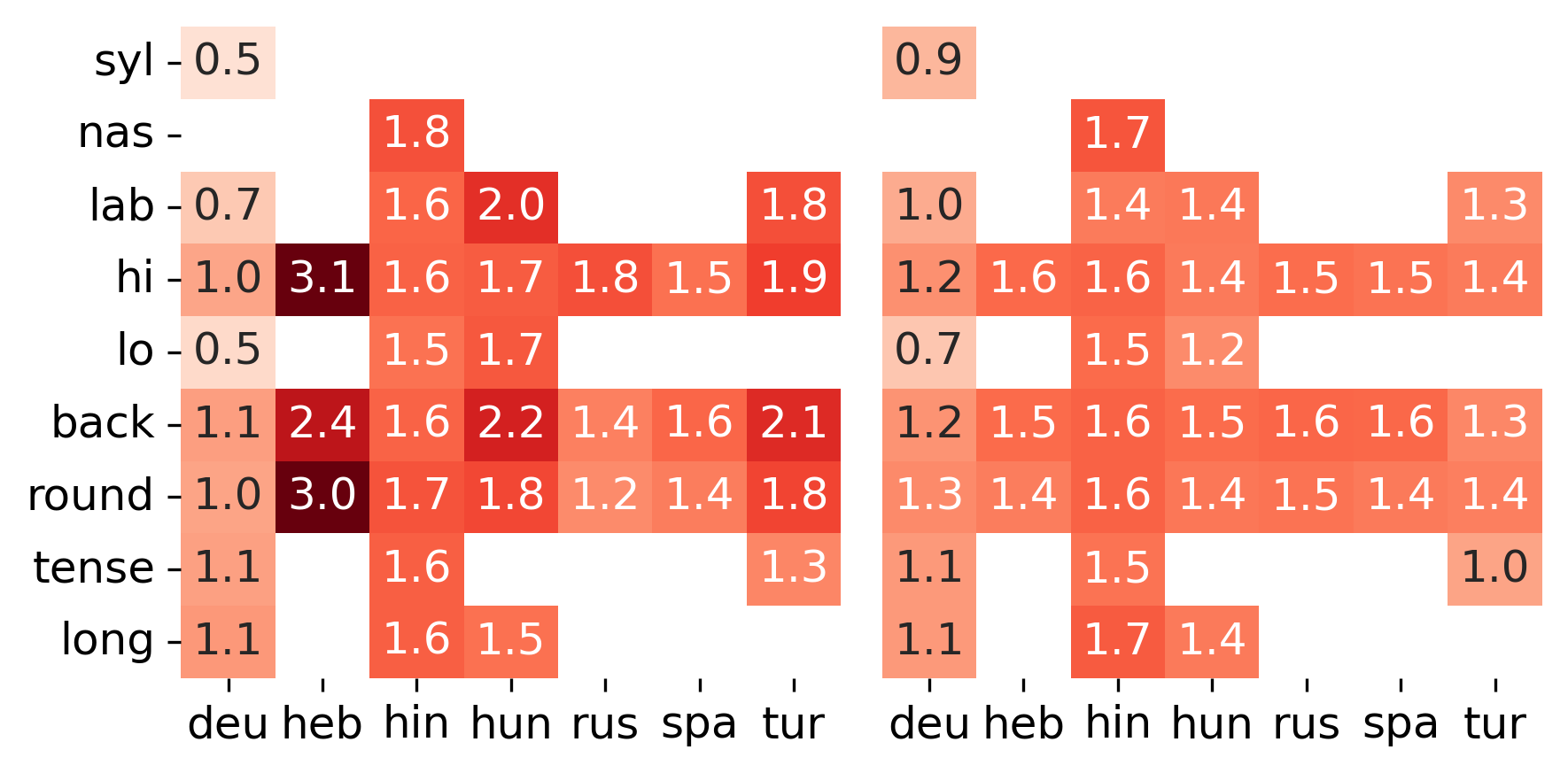}}
    \caption{Compression scores ($\mathcal{C}$) for probing vowel harmony. Inflection model on the left, lemma copying on the right.}
    \label{fig:vowel_inflection}
\end{figure}

\paragraph{Long-context feature agreement.}
Vowel harmony is a rule requiring all vowels in a word to share a specific phonological feature.
For example, Turkish and Hungarian exhibit vowel harmony for \texttt{ROUND} and \texttt{BACK}.
Since this rule influences both phonotactics and morphology, we expect these features to have high compression scores.
While this is not observed in the phoneme probe (\autoref{fig:inflection_experiment}), the harmony probe results (\autoref{fig:vowel_inflection}) show high compression scores for context-dependent embeddings in the inflection model.
Results for probing consonant harmony are provided in \autoref{sec:appendix}.

\section{Discussion} \label{sec:discussion}
We showed that a morphological transformer can effectively acquire phonological features.
The quality of their representation, as reflected by the compression score ($\mathcal{C}$), varies across features and languages, and is influenced by how informative they are per language.
Features prominent in phonotactics or in short-context environments are represented better in the phoneme embeddings, while those more present across long contexts are represented better through the encoder.
Higher scores are generally observed for features that are more central to morphology and phonology, though these results may also be influenced by the quality of the datasets or grapheme-to-phoneme tools.
Surprisingly, in the phoneme probe, the lemma copying model achieves on par with or even better than the inflection model.
We believe this might be due to dataset noise, as explored in prior work~\cite{wiemerslage-etal-2023-investigation}.
Future work could investigate the variance across languages and models.

Our findings complement work that showed that adding subphonemic features hardly improves model performance, suggesting these are already present in their representations~\cite{wiemerslage-etal-2018-phonological,guriel-etal-2023-morphological}.
Our lemma copying findings reinforce the common practice of pre-training models for this task before turning to inflection~\cite{yang-etal-2022-generalizing,liu-hulden-2022-transformer,anastasopoulos-neubig-2019-pushing}, which has been argued to succeed due to inducing \say{copy bias} and to coaxing attention modules towards monotonicity~\cite{aharoni-goldberg-2017-morphological}.
Finally, our results imply that the demonstrated success of transfer learning in morphological inflection, even between typologically unrelated languages~\cite{mccarthy-etal-2019-sigmorphon,elsner-2021-transfers},
might stem from the model's ability to acquire subphonemic features, which are approximately universal~\cite{Mielke2008} and therefore transferable.

\section{Conclusion}
\label{src:conclusion}
In this paper we analyze phonological transformers trained on a type-level morphological task, finding that these models acquire subphonemic features.
We show that the degree to which these features are embedded in the transformer's representation depends on the feature's importance for the morphology and phonology of the language it is trained for; and on the locality of the feature's importance: in the encoder, long-context features are more salient.

We use these results to explain empirical training methods used in the morphology inflection domain.
We hope this analysis will add an analytical tool in explaining morphological models using phonology acquisition.

\section*{Limitations} \label{sec:limitations}
Our suggested probing method that outputs a compression score, although language-agnostic, might have underlying biases in its components: the transliteration tools, the character-based transformer and the morphology inflection datasets' quality.
While discussing common strategies in morphology inflection, we omitted a popular data augmentation called \emph{data hallucination }~\cite{anastasopoulos-neubig-2019-pushing}, where 
new training examples are synthesized from existing training examples by identifying a (possibly discontinuous) word
stem and replacing this with a random character
sequence. Since this augmentation might indulge phonologically invalid words, we decided not to incorporate it to our method and results.

\section*{Acknowledgments}
We thank Evyatar Cohen for his valuable guidance regarding Hebrew grapheme-to-phoneme conversion, and the anonymous reviewers for their helpful
feedback.
This research was supported in part by the Israel Science Foundation (grant No. 1166/23).

\bibliography{anthology,custom}
\bibliographystyle{acl_natbib}

\newpage

\appendix

\section{Consonant Harmony Results}
\label{sec:appendix}
\begin{figure*}[ht]
    \centering
    \includegraphics[width=0.95\textwidth]{{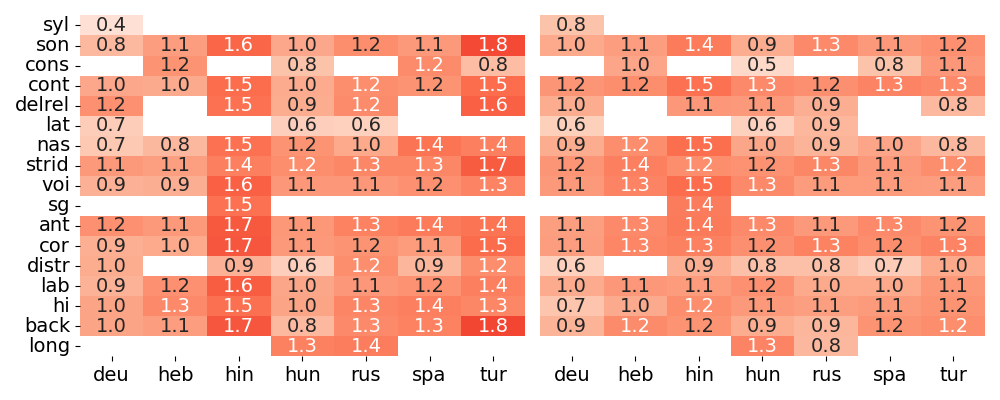}}
    \caption{Compression scores for probing consonant harmony. Inflection model on the left, lemma copying on the right.}
    \label{fig:consonant_inflection}
\end{figure*}

\autoref{fig:consonant_inflection} displays the results of the consonant harmony probe. 

\end{document}